\documentclass[sigconf]{acmart}

\usepackage[ruled]{algorithm2e} % For algorithms
\usepackage{algorithmic}

\usepackage{mathtools}

\usepackage{xcolor}
\usepackage{ifthen}
\newcommand{\markupdraft}[2]{% {#1: {color|display} command}{#2: desired color or text}
    \ifthenelse{\equal{#1}{display}}{#2}{}%                 % display only in draft version
    \ifthenelse{\equal{#1}{color}}{\color{#2}}{}%           % colored only in draft (for \new command)
}
\newcommand{\newcolored}[3][]{{\markupdraft{color}{#2}#3}%  % kept in the final print
    \ifthenelse{\equal{#1}{}}{}{\markupdraft{display}{{\color{yellow!70!black}[#1]}}}}
\newcommand{\del}[2][]{{\markupdraft{display}{{\color{orange}[removed: ``#2''[#1]]}}}} % (to be) removed
\newcommand{\new}[2][]{\newcolored[#1]{blue}{#2}}%  % kept in the final print
\newcommand{\nnew}[2][]{\newcolored[#1]{red}{#2}}%  % kept in the final print

\renewcommand{\del}[2]{}  % make removed sentences invisible
\renewcommand{\markupdraft}[2]{}  % remove all todo's, notes and coloring of changes
\newcommand{\rev}[1]{#1}
\newcommand{\revdel}[1]{}

\newtheorem{remark}{Remark}

\newcommand{\E}{\mathbb{E}}
\newcommand{\R}{\mathbb{R}}
\newcommand{\X}{\mathcal{X}}
\newcommand{\Z}{\mathcal{Z}}
\newcommand{\F}{\boldsymbol{F}}
\newcommand{\diff}{\mathrm{d}}
\newcommand{\Tr}{\mathrm{Tr}}
\newcommand{\I}{\mathbf{I}}
\newcommand{\T}{\mathrm{T}}
\newcommand{\N}{\mathcal{N}}
\newcommand{\argmax}{\mathop{\rm arg~max}\limits}
\newcommand{\argmin}{\mathop{\rm arg~min}\limits}

\newcommand{\btheta}{\boldsymbol{\theta}}
\newcommand{\bSigma}{\boldsymbol{\Sigma}}

\newcommand{\x}{\boldsymbol{x}}
\newcommand{\y}{\boldsymbol{y}}
\newcommand{\z}{\boldsymbol{z}}

\newcommand{\mv}{\boldsymbol{m}}
\newcommand{\cov}{\boldsymbol{C}}

\newcommand{\p}{\boldsymbol{p}}

\newcommand{\mvt}[1][t]{\boldsymbol{m}^{(#1)}}
\newcommand{\covt}[1][t]{\boldsymbol{C}^{(#1)}}
\newcommand{\stt}[1][t]{\sigma^{(#1)}}

\newcommand{\mueff}{\mu_{\mathrm{w}}}

\newcommand{\all}{{\mathrm{all}}}

\AtBeginDocument{%
  \providecommand\BibTeX{{%
    \normalfont B\kern-0.5em{\scshape i\kern-0.25em b}\kern-0.8em\TeX}}}

\copyrightyear{2024} 
\acmYear{2024} 
\setcopyright{acmlicensed}\acmConference[GECCO '24]{Genetic and Evolutionary Computation Conference}{July 14--18, 2024}{Melbourne, VIC, Australia}
\acmBooktitle{Genetic and Evolutionary Computation Conference (GECCO '24), July 14--18, 2024, Melbourne, VIC, Australia}
\acmDOI{10.1145/3638529.3654182}
\acmISBN{979-8-4007-0494-9/24/07}

\begin{document}

\title{CMA-ES with Adaptive Reevaluation for Multiplicative Noise}


\author{Kento Uchida}
\email{uchida-kento-fz@ynu.ac.jp}
\orcid{0000-0002-4179-6020}
\affiliation{%
  \institution{Yokohama National University}
  \city{Yokohama}
  \state{Kanagawa}
  \country{Japan}
  \postcode{240-8501}
}

\author{Kenta Nishihara}
\email{nishihara-kenta-xt@ynu.jp}
\orcid{0000-0003-4038-411X}
\affiliation{%
  \institution{Yokohama National University}
  \city{Yokohama}
  \state{Kanagawa}
  \country{Japan}
  \postcode{240-8501}
}

\author{Shinichi Shirakawa}
\email{shirakawa-shinichi-bg@ynu.ac.jp}
\orcid{0000-0002-4659-6108}
\affiliation{%
  \institution{Yokohama National University}
  \city{Yokohama}
  \state{Kanagawa}
  \country{Japan}
  \postcode{240-8501}
}

\renewcommand{\shortauthors}{K. Uchida et al.}

\begin{abstract}
% \note{196 / 200 words}
The covariance matrix adaptation evolution strategy (CMA-ES) is a powerful optimization method for continuous black-box optimization problems. Several noise-handling methods have been proposed to bring out the optimization performance of the CMA-ES on noisy objective functions. The adaptations of the population size and the learning rate are two major approaches that perform well under additive Gaussian noise. The reevaluation technique is another technique that evaluates each solution multiple times. In this paper, we discuss the difference between those methods from the perspective of stochastic relaxation that considers the maximization of the expected utility function. We derive that the set of maximizers of the noise-independent utility, which is used in the reevaluation technique, certainly contains the optimal solution, while the noise-dependent utility, which is used in the population size and leaning rate adaptations, does not satisfy it under multiplicative noise. Based on the discussion, we develop the reevaluation adaptation CMA-ES (RA-CMA-ES), which computes two update directions using half of the evaluations and adapts the number of reevaluations based on the estimated correlation of those two update directions. The numerical simulation shows that the RA-CMA-ES outperforms the comparative method under multiplicative noise, maintaining competitive performance under additive noise.
\end{abstract}




\maketitle

% --------------------------
\section{Introduction}
% --------------------------
The covariance matrix adaptation evolution strategy (CMA-ES)~\cite{hansen:1996:cmaes} is a probabilistic model-based evolutionary algorithm that employs a multivariate Gaussian distribution as the sampling distribution.
CMA-ES is known as a powerful optimization method for continuous black-box optimization.
The update procedure of CMA-ES consists of three steps: sampling the solutions from the multivariate Gaussian distribution, evaluating the solutions on the objective function, and updating the distribution parameters.
All the hyperparameters of the CMA-ES have \revdel{well-tuned}\rev{recommended} default values depending only on the search dimensions $d$ and population size $\lambda$~\cite{hansen:2011:tutorial}, which is given as $\lambda = 4 + \lfloor 3 \ln d \rfloor$ by default.
Based on the concept of stochastic relaxation~\cite{igo:2017}, the update rule of CMA-ES is partially considered as the natural gradient descent to maximize the expectation of the utility function~\cite{bidirectional}, and the ranking-based weights in CMA-ES are regarded as the estimated utility function values.

Objective functions often contain noise in real-world applications of evolutionary algorithms.
This noise disturbs the update direction of the distribution parameters and deteriorates the performance of the CMA-ES.
Several variants of the CMA-ES have been developed to improve the performance in noisy optimization problems.
The major categories of noise-handling methods for evolution strategies are categorized into three approaches: the reevaluations of solutions~\cite{Beyer:2007:resample, uhcmaes:2009}, population size adaptation~\cite{pccmsaes:2016, psacmaes:2018}, and learning rate adaptation~\cite{lracmaes:2023}.
Recently, adaptation methods of the population size~\cite{psacmaes:2018} and the learning rate~\cite{lracmaes:2023} for CMA-ES have been developed.
These methods maintain the signal-to-noise ratio (SNR) of the update directions of the distribution parameters and were evaluated using benchmark functions with additive Gaussian noise and demonstrated outstanding performance.
However, reevaluation adaptation, which adapts the number of evaluations for each solution, has not been actively investigated.

In this study, we discuss the difference between these noise-handling methods from the stochastic relaxation perspective.
We introduce two types of utility functions: {\it the noise-dependent utility}, which considers the maximization of the expectation under the joint distribution of the sampling and noise distributions, and {\it the noise-independent utility}, which is designed to maximize the expectation under the sampling distribution.
The population size and the learning rate adaptations employ the noise-dependent utility, whereas the reevaluation adaptation uses the noise-independent utility.
We also derive that the set of maximizers with noise-dependent utility does not contain the optimal solution certainly under multiplicative noise.
By contrast, the set of maximizers with noise-independent utility always contains the optimal solution under any type of noise.

We also propose a novel CMA-ES with a reevaluation adaptation mechanism, termed RA-CMA-ES.
Considering the success of adaptation using update directions, RA-CMA-ES computes two update directions using half of the evaluations and adapts the number of reevaluations based on the estimated correlation of those two update directions.
RA-CMA-ES also employs the learning rate adaptation to improve the estimation accuracy of the correlation.
We evaluated the performance of RA-CMA-ES on the benchmark functions under the additive noise and multiplicative noise.
Experimental results show that the RA-CMA-ES outperforms the comparative methods under multiplicative noise, while maintaining competitive performance under additive noise.
% \todo{simulation task}

% --------------------------
\section{Related Works}
% --------------------------
% --------------------------
\subsection{CMA-ES}
% --------------------------
The CMA-ES is a black-box optimization method for noiseless objective functions in $\R^d$.
The CMA-ES employs a multivariate Gaussian distribution
\begin{align}
    \N \left(\mvt, (\stt)^2 \covt \right) \enspace,
    \label{eq:cmaes:dist}
\end{align}
which is parameterized by the mean vector $\mvt \in \R^d$, covariance matrix $\covt \in \R^{d \times d}$, and step-size $\stt \in \R_{>0}$.
CMA-ES also has two evolution paths, $\p_\sigma^{(t)} \in \R^d$ and $\p_c^{(t)} \in \R^d$ which are initialized as $\p_\sigma^{(0)} = \p_c^{(0)} = \mathbf{0}$.

In each iteration, CMA-ES generates $\lambda$ solutions $\x_1, \cdots, \x_\lambda$ from the multivariate Gaussian distribution as
\begin{align}
    \z_i &\sim \N(\mathbf{0}, \I ) \\
    \y_i &= \sqrt{\covt} \z_i \\
    \x_i &= \mvt + \stt \y_i \enspace.
\end{align}
The CMA-ES then evaluates the solutions on the objective function and computes their rankings. In the following, we denote the index of $i$-th best solution as $i\!:\!\lambda$.
The CMA-ES assigns the weights $w_1, \cdots, w_\mu$ to the $\mu = \lfloor \lambda / 2 \rfloor$ best solutions as
\begin{align}
    w_i = \frac{\ln ((\lambda+1) / 2) - \ln i}{\sum_{k=1}^{\mu} \ln ((\lambda+1) / 2) - \ln k} \enspace.
    \label{eq:cma:weight}
\end{align}

Next, the CMA-ES updates two evolution paths using the weighted averages $\Delta_{\y} = \sum^{\mu}_{i=1} w_i \y_{i:\lambda}$ and $\Delta_{\z} = \sum^{\mu}_{i=1} w_i \z_{i:\lambda}$ as
\begin{align}
    \p_\sigma^{(t+1)} &= (1 - c_\sigma) \p_\sigma^{(t)} + \sqrt{c_\sigma (2 - c_\sigma) \mueff} \Delta_{\z} \\
    \p_c^{(t+1)} &= (1 - c_c) \p_c^{(t)} + h_\sigma^{(t+1)} \sqrt{c_c (2 - c_c) \mueff} \Delta_{\y} \enspace,
\end{align}
where $c_\sigma$ and $c_c$ are accumulation factors and $\mueff = (\sum^\mu_{i=1} w_i^2)^{-1}$ is the variance
effective selection mass and $h_\sigma^{(t+1)}$ is the Heaviside function.
We use the Heaviside function introduced in~\cite{principle:2014} as
\begin{align}
    h_\sigma^{(t+1)} = \begin{cases}
        1 & \text{if} \quad \frac{\| \p_\sigma^{(t+1)} \|^2}{1 - (1 - c_\sigma)^{2 (t+1)}} < \left( 2 + \frac{4}{d+1} \right) d \\
        0 & \text{otherwise} \enspace.
    \end{cases} 
\end{align}
Finally, CMA-ES updates the distribution parameters as
\begin{align}
    \label{eq:cmaes:mean-update}
    \mvt[t+1] &= \mvt + c_m \stt \Delta_{\y} \\
    \label{eq:cmaes:sigma-update}
    \stt[t+1] &= \stt \exp \left( \frac{c_\sigma}{d_\sigma} \left( \frac{\| \p_\sigma^{(t+1)} \| }{ \E[ \| \N(\mathbf{0}, \I ) \|] } - 1 \right) \right) \\
    \label{eq:cmaes:cov-update}
    \begin{split}
    \covt[t+1] &= \left( 1 + c_1 \delta^{(t+1)} \right) \covt + c_1 \left( \p_c^{(t+1)} \left( \p_c^{(t+1)} \right)^\T - \covt \right) \\
    &\hspace{80pt} + c_\mu \sum^\mu_{i=1} w_i \left( \y_{i:\lambda} \y_{i:\lambda}^\T - \covt \right) \enspace,
    \end{split}
\end{align}
where $c_m$, $c_1$, and $c_{\mu}$ are the learning rates, $d_\sigma$ is the damping factor, and $\delta^{(t+1)} = (1 - h_\sigma^{(t+1)}) c_c (2 - c_c)$. 
The expectation $\E[ \| \N(\mathbf{0}, \I ) \|]$ is typically approximated by $\sqrt{d} (1 - \frac{1}{4d} + \frac{1}{21 d^2})$.

CMA-ES has recommended settings for each hyperparameter~\cite{hansen:2011:tutorial, principle:2014}, which makes the CMA-ES a pseudo hyperparameter-free optimization method.

% --------------------------
\subsection{Noise-Resilient Variants of CMA-ES}
% --------------------------
In this section, we introduce several variants of the CMA-ES efficient on noisy optimization problems.
We consider the minimization of the expectation of the noisy objective function $f: \R^d \times \Z \to \R$ that receives the solution $\x \in \R^d$ and the noise $\z \in \Z$ generated from $P_n(\z \mid \x)$, that is,
\begin{align}
    \x^\ast = \argmin_{\x \in \X} \enspace \E_{\z \mid \x} [ f(\x, \z) ] \enspace, 
    \label{eq:original-problem}
\end{align}
where we denote $\E_{\z \mid \x} = \E_{\z \sim P_n(\z \mid \x)}$ for short.

% --------------------------
\paragraph{Uncertainty Handling CMA-ES}
% --------------------------
Uncertainty handling CMA-ES (UH-CMA-ES)~\cite{uhcmaes:2008, uhcmaes:2009} is a variant of CMA-ES for noisy optimization problems.
UH-CMA-ES measures the uncertainty level to incorporate the noise handling.
For $\lambda_\mathrm{reev}$ solutions selected from $\lambda$ solutions randomly, UH-CMA-ES performs the evaluation process twice and obtains two evaluation values ${f}_{i,1}$ and ${f}_{i,2}$ for each solution $\x_i$.
The evaluation processes for the other $\lambda - \lambda_\mathrm{reev}$ solution are performed once, and the evaluation values are copied as ${f}_{i,1} = {f}_{i,2}$.
Subsequently, the UH-CMA-ES computes the uncertainty level using the ranking in the union set $\{ {f}_{i,1} \}_{i=1}^\lambda \cup \{ {f}_{i,2} \}_{i=1}^\lambda$.
Denoting the ranking of $f_{i,1}$ and $f_{i,2}$ as $r_{i,1}$ and $r_{i,2}$, respectively, the uncertainty level is computed using the rank change $\Delta_i = | r_{i,1} - r_{i,2} | - 1$ as
\begin{multline}
    s = \frac{1}{\lambda_\mathrm{reev}} \sum_{i \in I_\mathrm{reev}} \biggl( 2 \Delta_i  - \Delta^{\lim}_{\theta} \left( \del{r_2}{}\new{r_{i,2}} - \mathbb{I}\{ f_{i,2} > f_{i,1} \} \right) \\
     - \Delta^{\lim}_{\theta} \left( \del{r_1}{}\new{r_{i,1}} - \mathbb{I}\{ f_{i,1} > f_{i,2} \} \right)  \biggr)
\end{multline}
where $I_\mathrm{reev}$ denotes the set of the indices of the $\lambda_\mathrm{reev}$ solutions. The function $\Delta^{\lim}_{\theta}(r)$ denotes the $(\theta / 2)$-quantile of the possible values of the rank change\del{ $|1 - r|, |2 - r|, \cdots, |2 \lambda - 1 - r|$}{}\new{, that is, $(\theta / 2)$-quantile in the set $\{ |1 - r|, |2 - r|, \cdots, |2 \lambda - 1 - r| \}$}.

For noisy black-box optimization problem, the uncertainty level is often used to adapt the number of reevaluations $n_\mathrm{eval}$ for each solution.
In the evaluation step of solutions, UH-CMA-ES evaluates each solution $n_\mathrm{eval}$ times and uses the average evaluation value to compute the ranking of the solutions as
\begin{align}
    \bar{f}_i = \frac{1}{n_\mathrm{eval}} \sum_{j=1}^{n_\mathrm{eval}} f(\x_i, \z_{i,j}) 
    \enspace.
    \label{eq:f:average}
\end{align}
The number of reevaluations is increased by a factor $\alpha$ when the uncertainty level is positive, and it is decreased by a factor $1 / \alpha$ otherwise.
See the detailed settings in~\cite{uhcmaes:2009}.

% --------------------------
\paragraph{Population Size Adaptation CMA-ES}
% --------------------------
Population size adaptation CMA-ES (PSA-CMA-ES)~\cite{psacmaes:2018} is a variant of CMA-ES with population size adaptation mechanism.
PSA-CMA-ES defines two update directions for the mean vector $\mv$ and the covariance $\boldsymbol{\Sigma} = \sigma^2 \cov$ of the sampling distribution as
\begin{align}
    \Delta_{\mv}^{(t+1)} &= \mv^{(t+1)}_\mathrm{ori} - \mv^{(t)} 
    \label{eq:sampling:m:diff} \\
    \Delta_{\bSigma}^{(t+1)} &= \mathrm{vec}\left( (\sigma^{(t+1)}_\mathrm{ori})^2 \cov^{(t+1)}_\mathrm{ori} - (\sigma^{(t)})^2 \cov^{(t)} \right) \enspace,
    \label{eq:sampling:Sigma:diff}
\end{align}
where $\mv^{(t+1)}_\mathrm{ori}$, $\sigma^{(t+1)}_\mathrm{ori}$, and $\cov^{(t+1)}_\mathrm{ori}$ are the updated distribution parameters in \eqref{eq:cmaes:mean-update}, \eqref{eq:cmaes:sigma-update}, and \eqref{eq:cmaes:cov-update}.%
\footnote{\new{
The operation $\mathrm{vec}(\cdot)$ transforms a given matrix $\boldsymbol{A} \in \R^{d \times d}$ to $d^2$-dimensional vector as $(\boldsymbol{A}_{1,1}, \cdots, \boldsymbol{A}_{1,d}, \boldsymbol{A}_{2,1}, \cdots, \boldsymbol{A}_{2,d}, \boldsymbol{A}_{3,1}, \cdots, \boldsymbol{A}_{d,d})^\T$.
}}
PSA-CMA-ES updates the population size so as to maintain the accuracy of the natural gradient estimation at a fixed target level.
PSA-CMA-ES introduces another evolution path $\p_\theta$ for each update direction $\Delta_{\btheta} = (\Delta_{\mv}^\T, \Delta_{\bSigma}^\T)^\T$ and its normalization factor $\gamma_\theta \in \R$ as
\begin{align}
    \p_{\theta}^{(t+1)} &= (1 - \beta) \p_{\theta}^{(t)} + \sqrt{2 (2 - \beta) } \frac{ \boldsymbol{F}_{\btheta^{(t)}}^{\frac12} \Delta_{\btheta}^{(t+1)} }{ \E \left[ \boldsymbol{F}_{\btheta^{(t)}}^{\frac12} \Delta_{\btheta}^{(t+1)} \right] } 
    \label{eq:psa:evopath} \\
    \gamma_\theta^{(t+1)} &= (1 - \beta)^2 \gamma_{\theta}^{(t)} + \beta (2 - \beta) \enspace,
\end{align}
where $\beta = 0.4$ is accumulation factor and $\boldsymbol{F}_{\btheta^{(t)}}$ is the Fisher information matrix of the sampling distribution.
The expectation in \eqref{eq:psa:evopath} is the expected norm under random selection, and the authors provide an approximation value.
PSA-CMA-ES updates the population size $\lambda^{(t)}$ as
\begin{align}
    \lambda^{(t + 1)} &= \lambda^{(t)} \exp \left( \beta \left( \gamma_\theta^{(t+1)} - \frac{ \| \p_{\theta}^{(t+1)} \|^2 }{ \alpha } \right) \right) \\
    \lambda^{(t + 1)} &\leftarrow \min\{ \max\{ \lambda^{(t + 1)}, \lambda_{\min} \}, \lambda_{\max} \} \enspace,
\end{align}
where $\alpha = 1.4$ and $\lambda_{\min}$ and $\lambda_{\max}$ are the target value of the norm of the \del{evaluation}{}\new{evolution} path and the minimum and the maximum population sizes, respectively.
After adapting the population size, the PSA-CMA-ES corrects the step-size using the optimal standard deviation derived from quality gain analysis~\cite{akimoto:2020:tcs}.

% --------------------------
\paragraph{Learning Rate Adaptation CMA-ES}
% --------------------------
Learning rate adaptation CMA-ES (LRA-CMA-ES)~\cite{lracmaes:2023} is a variant of CMA-ES that introduces a learning rate adaptation mechanism for the mean vector $\mv$ and covariance $\boldsymbol{\Sigma} = \sigma^2 \cov$ of the sampling distribution.
LRA-CMA-ES maintains the SNRs of those update directions in \eqref{eq:sampling:m:diff} and \eqref{eq:sampling:Sigma:diff} by updating the learning rates. 
To estimate the SNRs, LRA-CMA-ES computes the update directions $\tilde{\Delta}_{\btheta} = \sqrt{ \F_{\btheta} } \Delta_{\btheta}$ on the local coordinate system for $\btheta \in \{\mv, \bSigma\}$, which are given by
% Considering the squared roots of Fisher information matrices for $\mv$ and $\bSigma$, which are given by $\sqrt{\F_{\mv}} = \sqrt{\bSigma}^{-1}$ and $\sqrt{\F_{\bSigma}} = 2^{- \frac{1}{2}} \sqrt{\bSigma}^{-1} \otimes \sqrt{\bSigma}^{-1}$, we have
\begin{align}
    \tilde{\Delta}^{(t+1)}_{\mv} &=  \left( \bSigma^{(t)} \right)^{- \frac{1}{2}} \Delta_{\mv}^{(t+1)} 
    \label{eq:tdelta:m}\\
    \tilde{\Delta}^{(t+1)}_{\bSigma} &= \frac{1}{\sqrt{2}} \mathrm{vec}\left( \left( \bSigma^{(t)} \right)^{- \frac{1}{2}} \mathrm{vec}^{-1} \left( \Delta_{\bSigma}^{(t+1)} \right) \left( \bSigma^{(t)} \right)^{- \frac{1}{2}} \right) \enspace.
    \label{eq:tdelta:cov}
\end{align}
The LRA-CMA-ES also introduces two accumulations as
\begin{align}
    \label{eq:lra:acc-e}
    \mathcal{E}^{(t+1)}_{\btheta} &= (1 - \beta_{\btheta}) \mathcal{E}^{(t)}_{\btheta} + \beta_{\btheta} \tilde{\Delta}^{(t+1)}_{\btheta} \\
    \label{eq:lra:acc-v}
    \mathcal{V}^{(t+1)}_{\btheta} &= (1 - \beta_{\btheta}) \mathcal{V}^{(t)}_{\btheta} + \beta_{\btheta} \| \tilde{\Delta}^{(t+1)}_{\btheta} \|^2  \enspace,
\end{align}
where $\beta_{\btheta}$ denotes the accumulation factor.
Assuming that the distribution does not change significantly for some iterations, the LRA-CMA-ES estimates the SNR as
\begin{align}
    \frac{\| \E[ \tilde{\Delta}^{(t+1)}_{\btheta} ] \|^2}{\Tr (\mathrm{Cov}[ \tilde{\Delta}^{(t+1)}_{\btheta} ])} \approx \frac{ \| \mathcal{E}^{(t+1)}_{\btheta} \|^2 - \frac{\beta_{\btheta}}{2 - \beta_{\btheta}} \mathcal{V}^{(t+1)}_{\btheta} }{\mathcal{V}^{(t+1)}_{\btheta} - \| \mathcal{E}^{(t+1)}_{\btheta} \|^2} := \widehat{\mathrm{SNR}} \enspace.
\end{align}
LRA-CMA-ES updates the learning rate $\eta_{\btheta}$ such that the SNR is maintained at approximately $\alpha \eta_{\btheta}$ as
\begin{align}
    \eta_{\btheta}^{(t+1)} = \eta_{\btheta}^{(t)} \exp \left( \min \left\{ \gamma \eta_{\btheta}^{(t)}, \beta_{\btheta} \right\} \Pi_{[-1, 1]} \left( \frac{\widehat{\mathrm{SNR}}}{\alpha \eta_{\btheta}^{(t)}} - 1 \right) \right) \enspace,
    \label{eq:lra:update}
\end{align}
where $\Pi_{[-1, 1]}$ is the projection onto $[-1, 1]$, and $\alpha = 1.4$ and $\gamma = 0.1$ are hyperparameters. The accumulation factor is set $\beta_{\mv} = 0.1$ in the update for $\mv$ and $\beta_{\bSigma} = 0.03$ in the update for $\bSigma$. 
Subsequently, the distribution parameters are updated as
\begin{align}
    \mvt[t+1] &= \mvt + \eta_{\mv}^{(t+1)} \Delta_{\mv}^{(t+1)} 
    \label{eq:lra:update:m} \\
    \bSigma^{(t+1)} &= \bSigma^{(t)} + \eta_{\bSigma}^{(t+1)} \Delta_{\bSigma}^{(t+1)} 
    \label{eq:lra:update:bSigma} \\
    \stt[t+1] &= \det ( \bSigma^{(t+1)} )^{\frac{1}{2d}} 
    \label{eq:lra:update:sigma} \\
    \covt[t+1] &= \bSigma^{(t+1)} / (\stt[t+1])^2 
    \label{eq:lra:update:cov} \enspace.
\end{align}
Like PSA-CMA-ES, LRA-CMA-ES corrects the step-size $\stt[t+1]$ by multiplying ${\eta_{\mv}^{(t+1)}} / {\eta_{\mv}^{(t)}}$ such that the step-size is proportional to $\eta_{\mv}^{(t+1)}$.

% --------------------------
\section{Stochastic Relaxation in Noisy Optimization}
% --------------------------
Based on stochastic relaxation~\cite{igo:2017}, we can reformulate the minimization problem in~\eqref{eq:original-problem} as
\begin{align}
    \btheta^\ast = \argmin_{\btheta \in \Theta} \enspace \E_{\x} \left[ \E_{\z\mid\x}\left[ f(\x, \z) \right] \right] \enspace,
    \label{eq:sr-problem} 
\end{align}
where $\E_{\x} = \E_{\x \sim P_{\btheta}(\x)}$. For CMA-ES, the distribution $P_{\btheta}$ is the multivariate Gaussian distribution in \eqref{eq:cmaes:dist}. 
Some update rules of the CMA-ES can be explained by the natural gradient decent~\cite{bidirectional} as
\begin{align}
    \btheta^{(t+1)} = \btheta^{(t)} + \eta_{\btheta} \tilde{\nabla}_{\btheta} \E_{\x} \left[ \E_{\z\mid\x}\left[ f(\x, \z) \right] \right] \enspace.
\end{align}
Under weak assumptions, the optimal distribution $P_{\btheta^\ast}$ is given by the delta distribution around the optimal solution $\x^\ast$ \del{on $f$}{}\nnew{in~\eqref{eq:original-problem}}. 

In stochastic relaxation, the ranking-based weights are considered as the estimated utilities of the solutions, which are determined by the utility function using their quantile with respect to (w.r.t.) the current distribution.
Stochastic relaxation derives the update rules for the distribution parameters to maximize the expected utility function value.
For CMA-ES in noisy optimization, there are two possible choices for the distribution: the Gaussian distribution $P_{\btheta}(\x)$ and the joint distribution $\hat{P}_{\btheta}(\x, \z) = P_{\btheta}(\x) P_n(\z \mid \x)$ of the Gaussian and noise distributions.
In this section, we explain their differences and discuss the preferable setting for noisy optimization.

% ------------------
\subsection{Possible Setting of Utility Functions}
% ------------------
We discuss two possible settings for utility function: the noise-dependent utility $v(\x,\z; \btheta)$ and the noise-independent utility $u(\x; \btheta)$.
The noise-dependent utility computes the quantile w.r.t. the joint distribution $\hat{P}_{\btheta}(\x, \z)$ whereas the noise-independent utility $u(\x; \btheta)$ computes the quantile w.r.t. the Gaussian distribution $P_{\btheta}(\x)$.

\paragraph{Noise-dependent utility:}
The noise-dependent utility is applied to the CMA-ES with the population size and learning rate adaptations. 
The noise-dependent utility has two definitions of the quantile on $f$ w.r.t. $\hat{P}_{\btheta}(\x, \z)$ as
\begin{align}
    q_v^{<}(\x, \z; \btheta) &= \Pr_{\x', \z' \sim \hat{P}_{\btheta}} \left( f(\x', \z') < f(\x, \z) \right) \\
    q_v^{\leq}(\x, \z; \btheta) &= \Pr_{\x', \z' \sim \hat{P}_{\btheta}} \left( f(\x', \z') \leq f(\x, \z) \right) \enspace.
\end{align}
Introducing a non-increasing function $w: [0,1] \to \R$, called the selection scheme, we obtain the utility function as
\begin{align}
v(\x,\z; \btheta) =
\begin{dcases}
    w(q^{<}_v(\x, \z; \btheta)) & \text{if} \enspace q^{<}_v = q^{\leq}_v \\
    \frac{1}{q^{\leq}_v  - q^{<}_v } \int_{q = q^{<}_v }^{\new{q = }q^{\leq}_v } w(q) \diff q & \text{otherwise} \enspace, 
\end{dcases}
\label{eq:utility:v}
\end{align}
where we denote $q^{<}_v(\x, \z; \btheta)$ and $q^{\leq}_v(\x, \z; \btheta)$ as $q^{<}_v$ and $q^{\leq}_v$ for short, respectively.
We note that the quantile on the black-box function $f$ cannot be computed analytically.
Practically, the utility values are estimated using the Monte Carlo approximation with $\lambda$ pairs of solutions and random vectors $\{(\x_i, \z_i)\}_{i=1}^\lambda$ using \eqref{eq:utility:v} replacing the quantiles $q^{<}_v$ and $q^{\leq}_v$ with the estimated quantiles 
\begin{align}
    \bar{q}^{<}_{v}(\x_i, \z_i) &= \frac{1}{\lambda} \sum^{\lambda}_{k=1} \mathbb{I}_{ \left\{ f(\x_k, \z_k) < f(\x_i, \z_i) \right\}} \\
    \bar{q}^{\leq}_{v}(\x_i, \z_i) &= \frac{1}{\lambda} \sum^{\lambda}_{k=1} \mathbb{I}_{ \left\{ f(\x_k, \z_k) \leq f(\x_i, \z_i) \right\}} \enspace,
\end{align}
respectively.
\new{
This utility can recover the weights~\eqref{eq:cma:weight} used in the CMA-ES with corresponding selection scheme $w$.
}

\paragraph{Noise-independent utility:}
The noise-independent utility is the formulation for CMA-ES with reevaluation mechanism such as UH-CMA-ES. The quantiles on $f$ w.r.t. $P_{\btheta}$ are defined as
\begin{align}
    q_u^{<}(\x; \btheta) &= \Pr \left( \E_{\z \mid \x'}[ f(\x', \z) ] < \E_{\z \mid \x}[ f(\x, \z) ] \right) \\ 
    q_u^{\leq}(\x; \btheta) &= \Pr \left( \E_{\z \mid \x'}[ f(\x', \z) ] \leq \E_{\z \mid \x}[ f(\x, \z) ] \right) \enspace.
\end{align}
The noise-independent utility function is then given by 
\begin{align}
    u(\x; \btheta) = 
    \begin{dcases}
        w(q^{<}_u ) & \text{if} \enspace q^{<}_u  = q^{\leq}_u  \\
        \frac{1}{q^{\leq}_u  - q^{<}_u } \int_{q = q^{<}_u }^{\new{q = }q^{\leq}_u } w(q) \diff q & \text{otherwise} \enspace,
    \end{dcases}
    \label{eq:utility:u}
\end{align}
where we denote $q^{<}_u(\x, \z; \btheta)$ and $q^{\leq}_u(\x, \z; \btheta)$ as $q^{<}_u$ and $q^{\leq}_u$ for short, respectively.
The utilities are estimated using the Monte Carlo approximation with the evaluation values obtained by $n_\mathrm{eval}$ reevaluations $\{ f(\x_i, \z_{i,j}) \}_{j=1}^{n_\mathrm{eval}}$ for each of $\lambda$ solutions $\x_1, \cdots, \x_\lambda$.
The estimated utilities are given by~\eqref{eq:utility:u} replacing the quantiles $q^{<}_u$ and $q^{\leq}_u$ with the estimated quantiles
\begin{align}
    \bar{q}^{<}_{u}(\x_i) = \frac{1}{\lambda} \sum^{\lambda}_{k=1} \mathbb{I}_{ \left\{ \bar{f}_k < \bar{f}_i \right\}} \quad \text{and} \quad
    \bar{q}^{\leq}_{u}(\x_i) = \frac{1}{\lambda} \sum^{\lambda}_{k=1} \mathbb{I}_{ \left\{ \bar{f}_k \leq \bar{f}_i \right\}} \enspace,
\end{align}
where $\bar{f}_i$ denotes the average evaluation value in the reevaluations defined in \eqref{eq:f:average}, respectively.

% ------------------
\subsection{Discussion on Preferable Utility Function}
\label{sec:utility:discussion}
% ------------------
The natural gradient descents using these utilities maximize their expectations, 
which \new{are expected to} result in the distributions on the sets
\footnote{
We implicitly assume the existence of the expectations $\E_{z \mid \x}[ f(\x, z) ]$ and $\E_{z \mid \x}[ v(\x, z; \btheta) ]$.
}
\begin{align}
    \argmax_{\x \in \mathcal{X}} \enspace \E_{z \mid \x}[ v(\x, z; \btheta) ]
    \quad \text{and} \quad
    \argmax_{\x \in \mathcal{X}} \enspace u(\x; \btheta) \enspace,
    \label{eq:utility:maximizer}
\end{align}
respectively.
In this section, we present three lemmas to discuss the preferable utility setting for noisy optimization by comparing the set of maximizers in \eqref{eq:utility:maximizer}.
The proofs of those lemmas are provided in the supplementary material.
First, we provide a lemma related to the noise-dependent utility.

\begin{lemma} \label{lemma:dependent-multiplicative}
Assume the following conditions:
\begin{itemize}
    \item For any $\btheta \in \Theta$ and any set $S$ of a finite elements in $\X$, it holds $\Pr_{\x \sim P_{\btheta}}( \x \in S ) = 0$.
    \item The noise distribution generates scalar noise $z \in \R$ with positive expectation $\E[z] > 0$.
    \item The selection scheme $w$ is a strictly convex or strictly concave function.
\end{itemize}
Then, for any distribution parameter $\btheta \in \Theta$, there exist a function $f$ and a noise distribution $P_{n}$ that hold
\begin{align}
    \x^\ast \notin \argmax_{\x \in \mathcal{X}} \enspace \E_{z \mid \x}[ v(\x, z; \btheta) ] \enspace,
    %\label{eq:dependent-multiplicative-opt}
\end{align}
\nnew{where $\x^\ast$ is an arbitrary optimal solution in~\eqref{eq:original-problem}.}
\end{lemma}

\begin{remark} \label{remark:lemma1}
    The function $f$ in Lemma~\ref{lemma:dependent-multiplicative} is given in the form $f(\x, z) = z f_{\x}(\x) + b$ with some $f_{\x}: \R^d \to \R$ and $b \in \R$.
\end{remark}

Lemma~\ref{lemma:dependent-multiplicative} and Remark~\ref{remark:lemma1} show that maximizing the expected value of the noise-dependent utility may not agree with the minimizing the objective function with multiplicative noise.
This indicates that the performance of the PSA-CMA-ES and LRA-CMA-ES may be deteriorated by some type of noise, including the multiplicative noise.

In the next lemma, we also show a case in which the noise-dependent utility works efficiently.

\begin{lemma} \label{lemma:dependent-additive}
    Consider the objective function is given in the form $f(\x, z) = f_{\x}(\x) + z $ with some $f_{\x}: \R^d \to \R$ and noise $z \in \R$ generated from Gaussian distribution $\N(0, \sigma_n^2)$ with zero mean and standard deviation $\sigma_n > 0$.
    Then, for any function $f_{\x}$, distribution parameter $\btheta \in \Theta$, and selection scheme $w$, it holds
    \begin{align}
    \x^\ast \in \argmax_{\x \in \mathcal{X}} \enspace \E_{z \mid \x}[ v(\x, z; \btheta) ] \enspace,
    \label{eq:dependent-additive-opt}
\end{align}
\nnew{where $\x^\ast$ is an arbitrary optimal solution in~\eqref{eq:original-problem}.}
\end{lemma}

We note that several studies, including the references~\cite{psacmaes:2018, lracmaes:2023} of PSA-CMA-ES and LRA-CMA-ES, investigate the performance of their noise-handling methods under the additive Gaussian noise.
Lemma~\ref{lemma:dependent-additive} implies that the noise-dependent utility is preferable for such a situation.
However, considering Lemma~\ref{lemma:dependent-multiplicative}, the performance using the noise-dependent utility is unreliable for other types of noise.

Finally, we provide a lemma related to the noise-independent utility.

\begin{lemma} \label{lemma:independent}
For any objective function $f$ with a unique optimal solution $\nnew{\x^\ast = \argmin_{\x \in \X} \E_{\z \mid \x} [ f(\x, \z) ]}$, distribution parameter $\btheta \in \Theta$, and selection scheme $w$, it holds
\begin{align}
    \x^\ast \in \argmax_{\x \in \mathcal{X}} \enspace u(\x; \btheta) \enspace.
    \label{eq:independent-opt}
\end{align}
\end{lemma}

Lemma~\ref{lemma:independent} indicates that the set of maximizers of the noise-independent utility constantly contains the optimal solution.

% --------------------------
\section{Adaptive Reevaluation}
% --------------------------
Because the performance of CMA-ES using noise-independent utility is affected by the number of reevaluations $n_\mathrm{eval}$, an adaptation mechanism is necessary.
Considering the success of the CMA-ES with the adaptation mechanisms using the update directions of the distribution parameters, such as the PSA-CMA-ES and LRA-CMA-ES, 
% we develop an adaptation mechanism for the number of reevaluation $n_\mathrm{eval}$ using the correlation of two update directions $\Delta_{\btheta,1}$ and $\Delta_{\btheta,2}$ for $\btheta \in \{ \mv, \bSigma \}$ which are computed with the half of reevaluations.
we develop an adaptation mechanism for the number of reevaluation $n_\mathrm{eval}$ using the update directions for $\mv$ and $\bSigma$.
We compute two update directions $\Delta_{\btheta,1}$ and $\Delta_{\btheta,2}$ for $\btheta \in \{ \mv, \bSigma \}$ using different ranking of the average evaluation values over $n_\mathrm{half} = \lfloor n_\mathrm{eval} / 2 \rfloor$ reevaluations for the common solutions as
\begin{align}
    \bar{f}_{i,1} = \frac{\sum_{j=1}^{n_\mathrm{half}} f(\x_i, \z_{i,j}) }{n_\mathrm{half}} 
    \quad \text{and} \quad
    \bar{f}_{i,2} = \frac{\sum_{j=n_\mathrm{half}}^{2 n_\mathrm{half}} f(\x_i, \z_{i,j})}{n_\mathrm{half}}  
    \enspace.
    \label{eq:half:reeval}
\end{align}
we then estimate the correlation $\rho_{\btheta}$ of $\Delta_{\btheta,1}$ and $\Delta_{\btheta,2}$ and adapt the number of reevaluation to maintain a smaller correlation in $\rho_{\mv}$ and $\rho_{\bSigma}$ close to the target value.

% --------------------------
\subsection{Estimation of Update Correlation}
% --------------------------

We define the correlation between $\Delta_{\btheta,1}$ and $\Delta_{\btheta,2}$ as
\begin{align}
    \frac{ \Tr( \F_{\btheta} \mathrm{Cov}[ \Delta_{\btheta,1}, \Delta_{\btheta,2} ]) }{\sqrt{ \Tr( \F_{\btheta} \mathrm{Cov}[ \Delta_{\btheta,1} ])} \sqrt{\Tr( \F_{\btheta} \mathrm{Cov}[ \Delta_{\btheta,2} ]) }} \enspace,
\end{align}
where $\F_{\btheta}$ denotes the Fisher information matrix. 
% Our definition is the same as the Pearson correlation coefficient between $\tilde{\Delta}_{\btheta,1}$ and $\tilde{\Delta}_{\btheta,2}$, which are computed using \eqref{eq:tdelta:m} for $\Delta_{\mv}$ and \eqref{eq:tdelta:cov} for $\Delta_{\bSigma}$.
The concept of the adaptation mechanism is as follows: When the correlation is large, because the update direction \del{$\Delta_\all$ }{}is robust to noise, we decrease the number of reevaluations $n_\mathrm{eval}$ to reduce the number of evaluations. Otherwise, we increase $n_\mathrm{eval}$ to more precisely estimate the update direction on the noise-independent utility.

As we cannot obtain the correlation analytically, we introduce three accumulations to estimate the correlation as
\begin{align}
    \mathcal{E}_{\btheta, \ell}^{(t+1)} &= (1 - \beta_{\btheta}) \mathcal{E}_{\btheta, \ell}^{(t)} + \beta_{\btheta} \tilde{\Delta}_{\btheta, \ell}^{(t)} 
    \label{eq:E:accumulation} \\
    \mathcal{V}_{\btheta, \ell}^{(t+1)} &= (1 - \beta_{\btheta}) \mathcal{V}_{\btheta, \ell}^{(t)} + \beta_{\btheta} \| \tilde{\Delta}_{\btheta, \ell}^{(t)} \|^2 
    \label{eq:V:accumulation} \\
    \mathcal{I}_{\btheta}^{(t+1)} &= (1 - \beta_{\btheta}) \mathcal{I}_{\btheta}^{(t)} + \beta_{\btheta} (\tilde{\Delta}_{\btheta, 1}^{(t)})^\T \tilde{\Delta}_{\btheta, 2}^{(t)} \enspace,
    \label{eq:I:accumulation}
\end{align}
where $\ell \in \{1, 2\}$.
As well as LRA-CMA-ES, the accumulation factors are set as $\beta_{\mv} = 0.1$ and $\beta_{\bSigma} = 0.03$. 
We consider the case in which the distribution parameter stays around the same point for $n$ iterations, which allows us to regard the update directions $\Delta_{\btheta, \ell}^{(t + k)}$ for $k \in \{1, \cdots, n \}$ as independent and identically distributed (i.i.d.) samples from the same distribution with the expectation $\E[ \tilde{\Delta}_{\btheta, \ell} ]$ and covariance $\mathrm{Cov}[ \tilde{\Delta}_{\btheta, \ell} ]$. Subsequently, we have the expectations of $\| \mathcal{E}_\ell \|^2$ and $\mathcal{V}_\ell$ as
\begin{align}
    \lim_{n \to \infty} \E[ \| \mathcal{E}_{\btheta, \ell}^{(t+n)} \|^2 ] &= \| \E[ \tilde{\Delta}_{\btheta, \ell} ] \|^2 + \frac{\beta_{\btheta}}{2 - \beta_{\btheta}} \Tr( \mathrm{Cov}[ \tilde{\Delta}_{\btheta, \ell} ]) \\
    \lim_{n \to \infty} \E[ \mathcal{V}_{\btheta, \ell}^{(t+n)} ] &= \| \E[ \tilde{\Delta}_{\btheta, \ell} ] \|^2 + \Tr( \mathrm{Cov}[ \tilde{\Delta}_{\btheta, \ell} ])
\end{align}
Similarly, we have
\begin{align}
    \begin{split}
        &\lim_{n \to \infty} \E[ (\mathcal{E}_{\btheta, 1}^{(t+n)})^\T \F_{\btheta}^{(t)} \mathcal{E}_{\btheta, 2}^{(t+n)} ] \\
        &\hspace{30pt} = \E[ \tilde{\Delta}_{\btheta, 1} ]^\T \E[ \tilde{\Delta}_{\btheta, 2} ] + \frac{\beta_{\btheta}}{2 - \beta_{\btheta}} \Tr( \mathrm{Cov}[ \tilde{\Delta}_{\btheta, 1}, \tilde{\Delta}_{\btheta, 2} ]) 
    \end{split} \\
    &\lim_{n \to \infty} \E[\mathcal{I}_{\btheta}^{(t+n)}] = 
    \E[ \tilde{\Delta}_{\btheta, 1} ]^\T \E[ \tilde{\Delta}_{\btheta, 2} ] + \Tr( \mathrm{Cov}[ \tilde{\Delta}_{\btheta, 1}, \tilde{\Delta}_{\btheta, 2} ])
\end{align}
Finally, we estimate the correlation as
\begin{align}
    \rho^{(t+1)}_{\btheta} = \frac{\mathcal{I}_{\btheta}^{(t+1)} - (\mathcal{E}_1^{(t+1)})^\T \F_{\btheta}^{(t)} \mathcal{E}_2^{(t+1)}}{ \sqrt{ \left( \mathcal{V}_1^{(t+1)} - \| \mathcal{E}_1^{(t+1)} \|^2 \right) \left( \mathcal{V}_2^{(t+1)} - \| \mathcal{E}_2^{(t+1)} \|^2 \right) } } \enspace.
\end{align}

\begin{algorithm}[t] 
\caption{The Reevaluation Adaptation CMA-ES}
\begin{algorithmic}[1] \label{alg:racmaes}
\REQUIRE The objective function $f$ to be optimized
\REQUIRE $\mv^{(0)}, \cov^{(0)}, \sigma^{(0)}, n_\mathrm{min} = 1.2, n_\mathrm{eval} = 1.2, \rho_\mathrm{base} = 0.8$ %\COMMENT{initial distribution parameters}
\WHILE{termination conditions are not met}
\STATE Compute the number of reevaluations $\bar{n}_\mathrm{eval}$ as \eqref{eq:reev:det}.
\FOR{$i = 1$ to $\lambda$}
\STATE Generate $\y_i = ( \covt )^{\frac{1}{2}} \z_i$ with $\z_i \sim \N(\mathbf{0}, \I)$. 
\STATE Compute $\x_i = \mvt + \stt \y_i$.
\STATE Evaluate $f( \x_i )$ for $\bar{n}_\mathrm{eval}$ times.
\ENDFOR
\STATE Compute $\Delta_{\mv}$ and $\Delta_{\bSigma}$ with $\bar{n}_\mathrm{eval}$ reevaluations.
\IF{$\bar{n}_\mathrm{eval} \neq 1$}
\STATE Compute two update directions $\Delta_{\mv,1}, \Delta_{\bSigma,1}$ and \\$\Delta_{\mv,2}, \Delta_{\bSigma,2}$ with $\lfloor \bar{n}_\mathrm{eval} / 2 \rfloor$ reevaluations for $\mv$ and $\bSigma$, respectively.
\ELSE
\STATE Set $\Delta_{\mv,1} = \Delta_{\mv,2} = \Delta_{\mv}$ and $\Delta_{\bSigma,1} = \Delta_{\bSigma,2} = \Delta_{\bSigma}$.
\ENDIF
\STATE Update accumulations $\mathcal{E}_{\btheta,1}^{(t)}, \mathcal{E}_{\btheta,2}^{(t)}, \mathcal{V}_{\btheta,1}^{(t)}, \mathcal{V}_{\btheta,2}^{(t)}$ and $\mathcal{I}_{\btheta}^{(t)}$ using $\Delta_{\btheta,1}$ and $\Delta_{\btheta,2}$ as \eqref{eq:E:accumulation}, \eqref{eq:V:accumulation} and \eqref{eq:I:accumulation}.
\STATE Compute target correlation in \eqref{eq:target:corr}.
\STATE Update number of reevaluation $n_\mathrm{eval}$ as \eqref{eq:reev:update1} and \eqref{eq:reev:update2}.
\STATE Update the learning rates $\eta_{\mv}$ and $\eta_{\bSigma}$ as \eqref{eq:lra:update}.
\STATE Update distribution parameters $\mvt[t+1], \covt[t+1]$ and $\stt[t+1]$ using \eqref{eq:lra:update:m}, \eqref{eq:lra:update:sigma}, and \eqref{eq:lra:update:cov}, respectively.
\STATE $t \leftarrow t+1$
\ENDWHILE
\end{algorithmic} 
\end{algorithm}
\begin{table*}[t]
    \caption{List of benchmark functions used in our experiment}
    % \vspace{2mm}
    \label{table:benchmark-functions}
    \centering
    \begin{tabular}{l|l}
    \hline \text { Definitions } & \text { Initial Distribution Parameters } \\
    \hline \hline $f_{\text {Sphere }}(x)=\sum_{i=1}^d x_i^2$ & $m^{(0)}=[3, \ldots, 3], \sigma^{(0)}=2$ \\
    $f_{\text {Ellipsoid }}(x)=\sum_{i=1}^d\left(1000^{\frac{i-1}{d-1}} x_i\right)^2$ & $m^{(0)}=[3, \ldots, 3], \sigma^{(0)}=2$ \\
    $f_{\text {Rosenbrock }}(x)=\sum_{i=1}^{d-1}\left(100\left(x_{i+1}-x_i^2\right)^2+\left(x_i-1\right)^2\right)$ & $m^{(0)}=[0, \ldots, 0], \sigma^{(0)}=0.1$ \\
    $f_{\text {Ackley }}(x)=20-20 \cdot \exp \left(-0.2 \sqrt{\frac{1}{d} \sum_{i=1}^d x_i^2}\right)+e-\exp \left(\frac{1}{d} \sum_{i=1}^d \cos \left(2 \pi x_i\right)\right)$ & $m^{(0)}=[15.5, \ldots, 15.5], \sigma^{(0)}=14.5$ \\
    $f_{\text {Schaffer }}(x)=\sum_{i=1}^{d-1} (x_i^2+x_{i+1}^2 )^{0.25} \cdot\left[\sin ^2\left(50 \cdot (x_i^2+x_{i+1}^2 )^{0.1}\right)+1\right]$ & $m^{(0)}=[55, \ldots, 55], \sigma^{(0)}=45$ \\
    $f_{\text {Rastrigin }}(x)=10 d+\sum_{i=1}^d\left(x_i^2-10 \cos \left(2 \pi x_i\right)\right)$ & $m^{(0)}=[3, \ldots, 3], \sigma^{(0)}=2$ \\
    $f_{\text {Bohachevsky }}(x)=\sum_{i=1}^{d-1}\left(x_i^2+2 x_{i+1}^2-0.3 \cos \left(3 \pi x_i\right)-0.4 \cos \left(4 \pi x_{i+1}\right)+0.7\right)$ & $m^{(0)}=[8, \ldots, 8], \sigma^{(0)}=7$ \\
    $f_{\text {Griewank }}(x)=\frac{1}{4000} \sum_{i=1}^d x_i^2-\Pi_{i=1}^d \cos (x_i / \sqrt{i})+1$ & $m^{(0)}=[305, \ldots, 305], \sigma^{(0)}=295$ \\
    \hline
    \hline
\end{tabular}
\end{table*}
%
%
%

% --------------------------
\subsection{Proposed Method: RA-CMA-ES}
% --------------------------
We propose the reevaluation adaptation CMA-ES (RA-CMA-ES) using the estimation mechanism of the correlation.
Algorithm~\ref{alg:racmaes} shows the pseudocode of the RA-CMA-ES.
RA-CMA-ES introduces a relaxed number of reevaluations $n_\mathrm{eval}^{(t)} \in \R_{>0}$ and determines the number of reevaluations $\bar{n}_\mathrm{eval}^{(t)}$ stochastically as
\begin{align}
    \bar{n}_\mathrm{eval}^{(t)} = \begin{cases}
        \lfloor n_\mathrm{eval}^{(t)} \rfloor + 1 & \text{with probability} \enspace n_\mathrm{eval}^{(t)} - \lfloor n_\mathrm{eval}^{(t)} \rfloor \\
        \lfloor n_\mathrm{eval}^{(t)} \rfloor & \text{otherwise} \enspace.
    \end{cases}
    \label{eq:reev:det}
\end{align}
When $\bar{n}_\mathrm{eval}^{(t)} = 1$, which is preferable for noiseless optimization, RA-CMA-ES computes the update directions $\Delta_{\btheta, 1}$ and $\Delta_{\btheta, 2}$ using the same evaluation values. 
Otherwise, $\Delta_{\btheta, 1}$ and $\Delta_{\btheta, 2}$ are computed using the average evaluation values in $n_\mathrm{half} = \lfloor \bar{n}_\mathrm{eval}^{(t)} / 2 \rfloor$ reevaluations as \eqref{eq:half:reeval}.
Notably, the distribution parameters are updated using $\bar{n}_\mathrm{eval}^{(t)}$ reevaluations as in~\eqref{eq:f:average}.

Next, RA-CMA-ES computes the target correlation $\rho^{(t+1)}_\mathrm{target}$ using $n_\mathrm{eval}^{(t)}$.
Because the reevaluation mechanism requires many evaluations, the update with a low-accuracy update direction is \new{more} reasonable \new{than the additional increase of reevaluations} when $n_\mathrm{eval}^{(t)}$ is overly large.
Therefore, we set the target correlation $\rho^{(t+1)}_\mathrm{target}$ as 
\begin{align}
    \xi^{(t)} &= \left( 1 + \ln n_\mathrm{eval}^{(t)} - \ln n_\mathrm{min} \right) \cdot \min\left\{ n_\mathrm{eval}^{(t)} - 1, 1 \right\} \\
    \rho^{(t+1)}_\mathrm{target} &= (\rho_\mathrm{base})^{\xi^{(t)}} \label{eq:target:corr}
\end{align}
where $\rho_\mathrm{base} = 0.8$.
The first factor of $\xi^{(t)}$ prevents the unnecessary increase of the number of reevaluation, whereas the second factor corrects the \del{estimated value}{}\new{target correlation} when $n_\mathrm{eval}^{(t)} < 2$, where $\bar{n}_\mathrm{eval}^{(t)}$ may be set as $\bar{n}_\mathrm{eval}^{(t)} = 1$.

RA-CMA-ES adapts $n_\mathrm{eval}^{(t)}$ such that both correlations of the update directions for $\mv$ and $\bSigma$ are maintained above the target correlation.
The RA-CMA-ES computes a smaller correlation as
\begin{align}
    \rho^{(t+1)}_\mathrm{min} = \min \left\{ \rho^{(t+1)}_{\mv}, \rho^{(t+1)}_{\bSigma} \right\} \enspace.
\end{align}
RA-CMA-ES updates the number of reevaluation $n_\mathrm{eval}^{(t+1)}$ as
\begin{align}
    n_\mathrm{eval}^{(t+1)} &= n_\mathrm{eval}^{(t)} \exp \left( \gamma \Pi_{[-1,1]}\left( \frac{\rho^{(t+1)}_\mathrm{min} }{\rho^{(t+1)}_\mathrm{target} } - 1 \right) \right)
    \label{eq:reev:update1} \\
    n_\mathrm{eval}^{(t+1)} &\leftarrow \max \left\{ n_\mathrm{eval}^{(t+1)}, n_\mathrm{min} \right\} \enspace,
    \label{eq:reev:update2}
\end{align}
where we set $\gamma=0.1$ and $n_\mathrm{min} = 1.2$.

To improve the performance on multimodal problems, we incorporate the learning rate adaptation of the LRA-CMA-ES.
The estimation mechanism of the correlation regards the update direction in different iterations as i.i.d. samples, and the learning rate adaptation restricts the change of the distribution such that the SNR of the update direction is close to the target value.
Therefore, we consider the learning rate adaptation also improves the estimation accuracy of the correlation
% We consider the learning rate adaptation also improves the estimation accuracy of the correlation because our estimation regards the update direction in different iterations as i.i.d. samples and because the change of the distribution is restricted so that the SNR of the update direction is close to the target value.

%
%
%
\begin{figure*}[!t]
\centering
\includegraphics[width=0.99\linewidth]{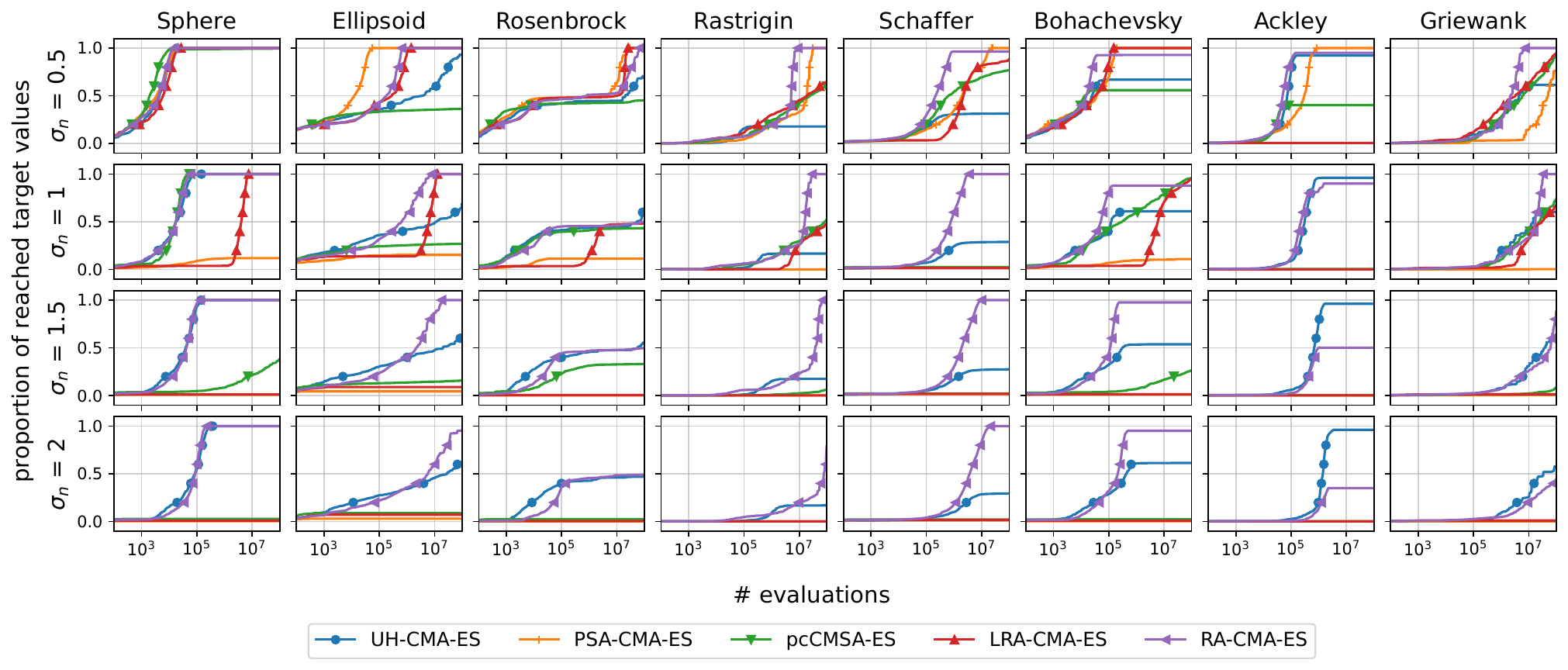}
\caption{Result with multiplicative Gaussian noise on 10-dimensional benchmark problems}
\label{fig:mul-gaussian}
\Description{This figure is the result with multiplicative Gaussian noise on 10-dimensional benchmark problems.}
\end{figure*}
\begin{figure*}[!t]
\centering
\includegraphics[width=0.99\linewidth]{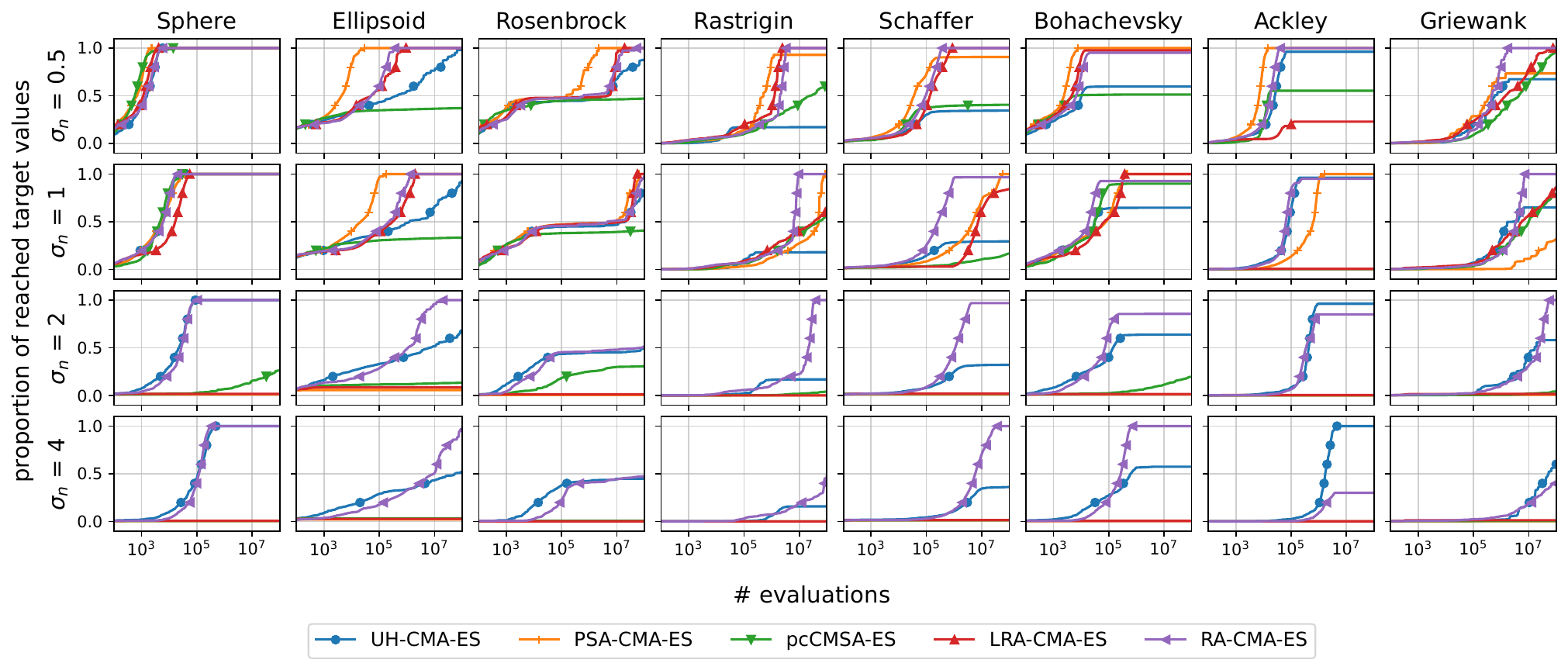}
\caption{Result with multiplicative uniform noise on 10-dimensional benchmark problems}
\label{fig:mul-uniform}
\Description{This figure is the result with multiplicative uniform noise on 10-dimensional benchmark problems.}
\end{figure*}
%
%
%

% --------------------------
\section{Experiment}
% --------------------------
We compare the optimization performance of the RA-CMA-ES with other variants of the CMA-ES on benchmark problems with additive noise and multiplicative noise.

% --------------------------
\subsection{Experimental Setting}
% --------------------------
We consider the three types of noisy objective functions.
\begin{itemize}
    \item Multiplicative Gaussian noise:
    \begin{align}
        f(\x, z) = f_{\x}(\x) \times (1 + \sigma_n z) \quad \text{where} \quad z \sim \N(0, 1)
    \end{align}
    \item Multiplicative uniform noise:
    \begin{align}
        f(\x, z) = f_{\x}(\x) \times (1 + \sigma_n z) \quad \text{where} \quad z \sim \mathrm{Uni}(-1, 1)
    \end{align}
    \item Additive Gaussian noise:
    \begin{align}
        f(\x, z) = f_{\x}(\x) + \sigma_n z \quad \text{where} \quad z \sim \N(0, 1)
    \end{align}
\end{itemize}
We note that $\mathrm{Uni}(a,b)$ denotes the uniform distribution on $[a,b]$.
We used the benchmark functions listed in Table~\ref{table:benchmark-functions} as the noiseless objective function $f_{\x}$, which was used in~\cite{lracmaes:2023}.
Table~\ref{table:benchmark-functions} also lists the initial mean vector and the initial step-size.
The initial covariance matrix is given by the identity matrix $\cov^{(0)} = \I$ for all the settings.
We set the number of dimensions to $d=10$.

We compare RA-CMA-ES with UH-CMA-ES, PSA-CMA-ES, and LRA-CMA-ES.
\footnote{
These implementations were used in this study.
\url{https://gist.github.com/youheiakimoto/f342a139a6dfcfd55dbc5e345dbf4f9f} for the PSA-CMA-ES and
\url{https://github.com/nomuramasahir0/cma-learning-rate-adaptation} for the LRA-CMA-ES.
}
In addition, we prepared pcCMSA-ES~\cite{pccmsaes:2016} for comparison.
The pcCMSA-ES controls the population size based on the estimated noise strength on the objective function.
We evaluated the optimization performance of each method using the empirical cumulative density function used in the COCO platform~\cite{coco}.
We introduced $N_\mathrm{target} = 500$ target evaluation values on the noiseless objective function $f_{\x}$ and recorded the proportion of target values reached by the evaluation value $f_{\x}(\mv^{(t)})$ at the mean vector in 20 independent trials.
We set the target evaluation values using points with even intervals on a log scale between $f_{\x}(\mv^{(0)})$ and $10^{-3}$.

\begin{figure*}[!t]
\centering
\includegraphics[width=0.99\linewidth]{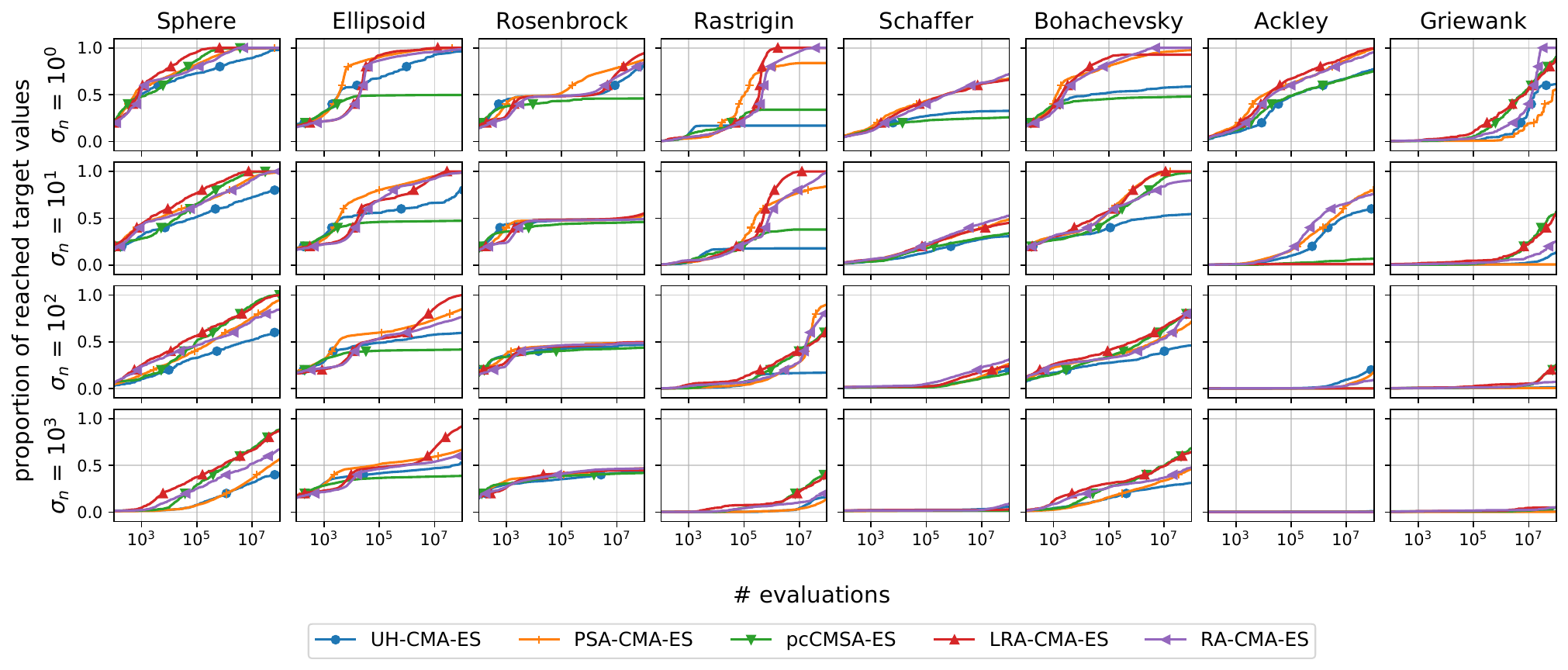}
\caption{Result with additive Gaussian noise on 10-dimensional benchmark problems}
\label{fig:add-gaussian}
\Description{This figure is the result with additive Gaussian noise on 10-dimensional benchmark problems.}
\end{figure*}
%
%
%

% --------------------------
\subsection{Result with Multiplicative Noise}
% --------------------------
Figures~\ref{fig:mul-gaussian} and \ref{fig:mul-uniform} show the results for multiplicative Gaussian and uniform noises, respectively.
We varied the noise strength as $\sigma_n = 0.5, 1, 1.5, 2$ for multiplicative Gaussian noise and $\sigma_n = 0.5, 1, 2, 4$ for multiplicative uniform noise.
We observed that RA-CMA-ES and UH-CMA-ES, which also uses a reevaluation technique, could improve the evaluation value at the mean vector with the strongest multiplicative noise.
Although the UH-CMA-ES achieved superior evaluation values compared to RA-CMA-ES in the early phase of the optimization, the RA-CMA-ES outperformed it at the end of the optimization on several functions.
The UH-CMA-ES uses the ranking information for the adaptation, whereas the RA-CMA-ES uses the accumulations of the update directions.
We contend that the update direction offers more insight than the ranking of the solution because the update directions computed with different rankings can be analogous.
On the Ackley function, however, RA-CMA-ES was significantly worse than the UH-CMA-ES.
We note that LRA-CMA-ES failed to optimize the Ackley function even under weak multiplicative noise.
We consider that the learning rate adaptation was not effective on the Ackley function with multiplicative noise, which makes a performance gap between the UH-CMA-ES and RA-CMA-ES.
Focusing on the results under weak noise $\sigma_n = 0.5$, the RA-CMA-ES worked better or competitive with the UH-CMA-ES.
We believe that UH-CMA-ES may increase the number of reevaluations than necessary.
% On the multimodal functions, such as the Rastrigin and Schaffer functions, we consider the leaning rate adaptation in the RA-CMA-ES improved the performance compared with the UH-CMA-ES.

% --------------------------
\subsection{Result with Additive Noise}
% --------------------------
Figure~\ref{fig:add-gaussian} shows the results for additive Gaussian noise.
We varied the noise strength as $\sigma_n = 10^0, 10^1, 10^2, 10^3$.
Compared with the UH-CMA-ES, the RA-CMA-ES outperformed the UH-CMA-ES with most of the experiment settings.
Focusing on the CMA-ES without reevaluation technique, we observe that the RA-CMA-ES was competitive with the PSA-CMA-ES and LRA-CMA-ES.
% We consider that the reevaluation adaptation worked well with the learning rate adaptation.
As discussed in Section~\ref{sec:utility:discussion}, we consider that the utility functions of PSA-CMA-ES and LRA-CMA-ES are preferable for additive Gaussian noise.
\rev{In contrast, the performance of RA-CMA-ES was powerful under both multiplicative and additive noises.}

% --------------------------
\section{Conclusion}
% --------------------------
In this study, we analyzed the existing noise-handling methods from the stochastic relaxation perspective.
We showed that the noise-dependent utility used in PSA-CMA-ES and LRA-CMA-ES is not suitable for some noisy optimization problems, particularly under multiplicative noise.
Subsequently, we focused on the reevaluation adaptation as a preferable approach for noisy optimization problems and proposed a novel reevaluation adaptation method for CMA-ES.
The proposed method, RA-CMA-ES, computes two update directions using half of the evaluations and adapts the number of reevaluations based on the estimated correlation of those two update directions.
RA-CMA-ES employs the learning rate adaptation to meet the assumption in the estimation mechanism of the correlation, thereby enhancing the correlation estimation accuracy.
We evaluated the performance of RA-CMA-ES on the benchmark functions under the additive noise and multiplicative noise.
Experimental results showed that the RA-CMA-ES outperformed the comparative method under multiplicative noise, while maintaining competitive performance under additive noise.

% \todo{future work}
% Because our adaptation mechanism contains some heuristic components, such as the adaptation of target correlation, we need to improve or justify those components in our future work.
Given that our adaptation mechanism incorporates heuristic elements like target correlation adaptation, we must refine or justify these components in future work.
In addition, the performance evaluation of the RA-CMA-ES with other types of noise, including the input noise and solution-dependent noise $P_n(\z \mid \x)$, is another topic for future work.

\rev{
% --------------------------
\section*{Acknowledgement}
% --------------------------
This work was partially supported by JSPS KAKENHI (JP23H00491, JP23H03466), JST PRESTO (JPMJPR2133), and NEDO (JPNP18002, JPNP20006).
}

\bibliographystyle{ACM-Reference-Format}
\bibliography{reference}

\clearpage

\appendix

% --------------------------
\section{Proof of Lemmas}
% --------------------------

% --------------------------
\subsection{Proof of Lemma~3.1}
\label{proof:lemma:dependent-multiplicative}
% --------------------------
\begin{proof}
Let us consider the objective function in the form $f(\x, z) = z f_{\x}(\x) + b$ with a constant value $b \in \R$.
We consider the case that the noise $z$ takes $1$ or $-1$ only, i.e. $z \in \{-1, 1\}$, and given independently from $\x$.
We will construct a function $f_{\x}$ satisfying 
\begin{align}
    \x^\ast \notin \argmax_{\x \in \mathcal{X}} \enspace \E_{z \mid \x}[ v(\x, z; \btheta) ] \enspace.
    %\label{eq:dependent-multiplicative-opt}
\end{align}
\nnew{for arbitrary optimal solution $\x^\ast$ in \eqref{eq:original-problem}} with convex and concave selection scheme $w$ separately.

\paragraph{Convex case:}
Consider the objective function is given \nnew{with a unique optimal solution $\x^\ast$ as}
\begin{align}
    f_{\x}(\x) = \begin{cases}
        \, 0 & \text{if} \enspace \x = \x^\ast \\
        \, 1 & \text{otherwise} \enspace.
    \end{cases}
\end{align} 
Considering $\Pr_{\x' \sim P_{\btheta}}( \x' \neq \x^\ast ) = 1$, the quantiles $q^{<}_v$ and $q^{\leq}_v$ are given by
\begin{align}
    & q^{<}_v(\x, z) = \begin{cases}
        \, p_{z}^{-} & \text{if} \enspace \x = \x^\ast \\
        \, 0 & \text{if} \enspace \x \neq \x^\ast \quad \text{and} \quad z = -1  \\
        \, p_{z}^{-} & \text{if} \enspace \x \neq \x^\ast \quad \text{and} \quad z = 1  
    \end{cases} \\
    & q^{\leq}_v(\x, z) = \begin{cases}
        \, p_{z}^{-} & \text{if} \enspace \x = \x^\ast \\
        \, p_{z}^{-} & \text{if} \enspace \x \neq \x^\ast \quad \text{and} \quad z = -1  \\
        \, 1 & \text{if} \enspace \x \neq \x^\ast \quad \text{and} \quad z = 1  
    \end{cases} \enspace.
\end{align} 
where we denote $p_{z}^{-} = \Pr_{z' \sim P_n}(z' = -1)$ for short.
The utility value is determined for $\x^\ast$ and $\x' \neq \x^\ast$ as
\begin{align}
    v(\x^\ast, z; \theta) &= w( p_{z}^{-} ) \\
    v(\x', -1; \theta) &= \frac{1}{p_{z}^{-}} \int_0^{p_{z}^{-}} w(q) \diff q \\
    v(\x', 1; \theta) &= \frac{1}{1 - p_{z}^{-}} \int_{p_{z}^{-}}^1 w(q) \diff q 
\end{align}
The expected utility is given by
\begin{align}
&\E_{\z \mid \x}[ v(\x, z; \theta) \mid \x = \x^\ast] = w(\Pr(z = -1)) \\
\begin{split}
    &\E_{\z \mid \x}[ v(\x, z; \theta) \mid \x \neq \x^\ast] \\
    & = \int_{0}^{p_{z}^{-}} w(q) \diff q + \int_{p_{z}^{-}}^{1} w(q) \diff q = \int_{0}^{1} w(q) \diff q \enspace.
\end{split}
\end{align}
Since $w$ is strictly convex, Jensen's inequality shows that there exists $\epsilon > 0$ that holds
\begin{align}
    \int_{0}^{1} w(q) \diff q = w \left( \frac{1}{2} \right) + \epsilon \enspace.
\end{align}
Moreover, we can consider arbitrary small $\delta \in (0, 1/2)$ which satisfies $p_{z}^{-} = \Pr(z = -1) = 1/2 - \delta$.
We have 
\begin{align}
    \begin{split}
        & \E_{\z \mid \x}[ v(\x, z; \theta) \mid \x \neq \x^\ast] - \E_{\z \mid \x}[ v(\x, z; \theta) \mid \x = \x^\ast] \\
        & \qquad = \int_{0}^{1} w(q) \diff q - w(p_{z}^{-}) 
    \end{split}\\
    & \qquad = \epsilon - \left( w \left( \frac{1}{2} - \delta \right) - w \left( \frac{1}{2} \right) \right)
\end{align}
Since $w$ is convex function on the open interval $(0, 1)$, it is continuous on $(0, 1)$. Therefore, for any $\epsilon$, there exists $\delta > 0$ which holds
\begin{align}
    \E_{\z \mid \x}[ v(\x, z; \theta) \mid \x \neq \x^\ast] > \E_{\z \mid \x}[ v(\x, z; \theta) \mid \x = \x^\ast] \enspace.
\end{align}

\paragraph{Concave case:}
Consider the objective function is given \nnew{with a unique optimal solution $\x^\ast$ as} 
\begin{align}
    f_{\x}(\x) = \begin{cases}
        \, -1 & \text{if} \enspace \x = \x^\ast \\
        \, 0 & \text{otherwise} 
    \end{cases} \enspace.
\end{align} 
In this case, the quantiles $q^{<}$ and $q^{\leq}$ are give by
\begin{align}
    & q^{<}(\x, z) = \begin{cases}
        \, 0 & \text{if} \enspace \x = \x^\ast \quad \text{and} \quad z = 1  \\
        \, 1 & \text{if} \enspace \x = \x^\ast \quad \text{and} \quad z = -1  \\
        \, 1 & \text{otherwise} 
    \end{cases} \\
    & q^{\leq}(\x, z) = \begin{cases}
        \, 0 & \text{if} \enspace \x = \x^\ast \quad \text{and} \quad z = 1  \\
        \, 1 & \text{if} \enspace \x = \x^\ast \quad \text{and} \quad z = -1  \\
        \, 0 & \text{otherwise} 
    \end{cases}\enspace.
\end{align} 
Then, the expected utility is given by
\begin{align}
\E_{\z \mid \x}[ v(\x, z; \theta) \mid \x = \x^\ast] &= \Pr(z = 1) w(0) + \Pr(z = -1) w(1) \\
\E_{\z \mid \x}[ v(\x, z; \theta) \mid \x \neq \x^\ast] &= \int_{0}^{1} w(q) \diff q \enspace.
\end{align}
For strictly concave function, we have
\begin{align}
    \int_{0}^{1} w(q) \diff q &> \int_{0}^{1} \left( (1-q) w(0) + q w(1) \right) \diff q \\
    &= \frac{3}{4} w(0) + \frac{1}{4} w(1) \enspace.
\end{align}
Therefore, setting $\Pr(z = 1) = 3/4$ satisfies
\begin{align}
    \E_{\z \mid \x}[ v(\x, z; \theta) \mid \x \neq \x^\ast] > \E_{\z \mid \x}[ v(\x, z; \theta) \mid \x = \x^\ast] \enspace.
\end{align}
This is end of the proof.
\end{proof}

% --------------------------
\subsection{Proof of Lemma~3.2}
\label{proof:lemma:dependent-additive}
% --------------------------
\begin{proof}
Because the probability of the event $f(\x,z) = f(\x', z')$ is zero for any $x$ and $x'$ under additive Gaussian noise, we have
\begin{align}
\E_{\z \mid \x}[ v(\x, z; \theta) ] &= \E_{\z \mid \x} \left[ w \left( q^{<}_v(\x, z; \theta) \right) \right] \\
&= \E_{\z \mid \x} \left[ w \left( \Pr_{\x', z' \sim \hat{P}_{\btheta}} \left( f(\x', z') < f(\x, z) \right) \right) \right]
\end{align}
Fixing the noise $\z$, an optimal solution $\x^\ast$ \nnew{in~\eqref{eq:original-problem}} satisfies $f(\x^\ast, z) \leq f(\x, z)$ for any $\x$, and we have
\begin{align*}
    \Pr_{\x', z' \sim \hat{P}_{\btheta}} \left( f(\x', z') < f(\x^\ast, z) \right) \leq \Pr_{\x', z' \sim \hat{P}_{\btheta}} \left( f(\x', z') < f(\x, z) \right)
\end{align*}
Finally, because the selection scheme $w$ is non-increasing function, it holds for any $\x$ as
\begin{align}
    \E_{\z \mid \x}[ v(\x^\ast, z; \theta) ] \geq \E_{\z \mid \x}[ v(\x, z; \theta) ]
\end{align}
This is end of the proof.
\end{proof}

% --------------------------
\subsection{Proof of Lemma~3.3}
\label{proof:lemma:independent}
% --------------------------
\begin{proof}
Because the optimal solution $\nnew{\x^\ast = \argmin_{\x \in \X} \E_{\z \mid \x} [ f(\x, \z) ]}$ is unique, we have $q_u^{<}(\x^\ast; \btheta) = q_u^{\leq}(\x^\ast; \btheta) = 0$ for any $\btheta \in \Theta$. Then, reminding the utility function $u( \, \cdot \,; \btheta)$ is non-increasing w.r.t. $q_u^{<}( \, \cdot \,; \btheta)$, the proof is finished.
\end{proof}

% insert the following to make it look good
\vspace{100pt}

\end{document}